\documentclass[letterpaper, 10 pt, conference]{ieeeconf}  

\IEEEoverridecommandlockouts                              

\usepackage{graphics} 
\usepackage{epsfig} 
\usepackage{mathptmx} 
\usepackage{times} 
\usepackage{amsmath} 
\usepackage{amssymb}  

\usepackage{amsfonts}
\usepackage{algorithmic}
\usepackage{graphicx}
\usepackage{textcomp}
\usepackage{xcolor}
\usepackage{kotex}
\usepackage{subfigure}
\usepackage{hyperref}
\usepackage{flushend}
\usepackage{balance}
\def\BibTeX{{\rm B\kern-.05em{\sc i\kern-.025em b}\kern-.08em
    T\kern-.1667em\lower.7ex\hbox{E}\kern-.125emX}}
    
\title{\LARGE \bf
Cyclops: Open Platform for Scale Truck Platooning
}

\author{Hyeongyu Lee$^{1}$, Jaegeun Park$^{2}$, Changjin Koo$^{1}$, Jong-Chan Kim$^{1}$, and Yongsoon Eun$^{2}$
\thanks{$^{1}$Hyeongyu Lee, Changjin Koo are graduate students and Jong-Chan Kim is a professor at the Graduate School of Automotive Engineering, Kookmin University, Seoul, Republic of Korea
        {\tt\small \{a2020011, cjkoo96, jongchank\}@kookmin.ac.kr.}}%
\thanks{$^{2}$Jaegeun Park is a graduate student and Yongsoon Eun is a professor at the Department of Information and Communication Engineering, DGIST, Daegu, Republic of Korea
        {\tt\small \{jaegeun2, yeun\}@dgist.ac.kr.}}%
}

\begin{document}

\maketitle
\thispagestyle{empty}
\pagestyle{empty}

\begin{abstract}

Cyclops, introduced in this paper, is an open research platform for everyone who wants to validate novel ideas and approaches in self-driving heavy-duty vehicle platooning. The platform consists of multiple 1/14 scale semi-trailer trucks equipped with associated computing, communication and control modules that enable self-driving on our scale proving ground. The perception system for each vehicle is composed of a lidar-based object tracking system and a lane detection/control system. The former maintains the gap to the leading vehicle, and the latter maintains the vehicle within the lane by steering control. The lane detection system is optimized for truck platooning, where the field of view of the front-facing camera is severely limited due to a small gap to the leading vehicle. This platform is particularly amenable to validating mitigation strategies for safety-critical situations. Indeed, the simplex architecture is adopted in the computing modules, enabling various fail-safe operations. In particular, we illustrate a scenario where the camera sensor fails in the perception system, but the vehicle is able to operate at a reduced capacity to a graceful stop. 
Details of Cyclops, including 3D CAD designs and algorithm source codes, are released for those who want to build similar testbeds.


\end{abstract}

\section{INTRODUCTION}

Vehicle platooning is an automation technology that coordinates multiple vehicles so that short longitudinal inter-vehicle distances are maintained while vehicles are running at a high speed~\cite{bergenhem2012overview}. The {\em leading vehicle} (LV) is usually driven by a human driver, while the {\em following vehicles} (FVs) are autonomous based on perception sensors and wireless communications between vehicles~\cite{bergenhem2012vehicle}. Its potential benefits include (i) reduced fuel consumption and emission, (ii) better road utilization, and (iii) enhanced safety~\cite{tsugawa2016review}~\cite{lioris2016doubling}.

There have been many national-level projects for vehicle platooning, including SARTRE~\cite{robinson2010operating}~\cite{bergenhem2010challenges} and ENSEMBLE~\cite{konstantinopoulou2019specifications} in EU, and TROOP~\cite{lee2020novel} in Korea. We are partly engaged in one of such projects that develops safety scenarios for heavy-duty truck platooning, where the developed system should be validated with potentially fatal scenarios on public highways. The scenarios include (i) emergency braking, (ii) cutting-in rogue vehicles, and (iii) loss of wireless connections. Due to the heavyweights of the employed semi-trailer trucks, we found it much too dangerous to put full-size trucks directly into such scenarios in early validation stages. Although Hardware-in-the-Loop (HiL) simulators are employed, they are mostly for validating control algorithms, not for validating countermeasures for safety-critical scenarios.

\begin{figure}
\vspace{0.285cm}
\hspace{0.05 cm}
\centerline{\includegraphics[width=7.75cm]{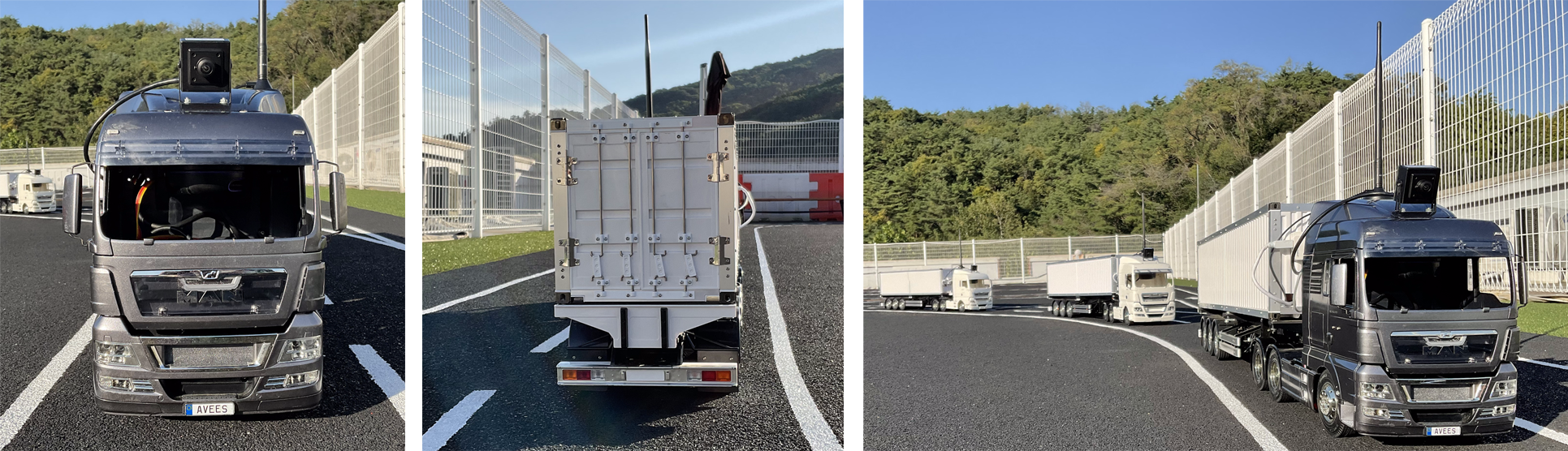}}
\vspace{-0.5cm}
\caption{Scale semi-trailer trucks in Cyclops.}
\label{fig:testbed}
\end{figure}

\begin{figure}
\centerline{\includegraphics[width=8.0cm]{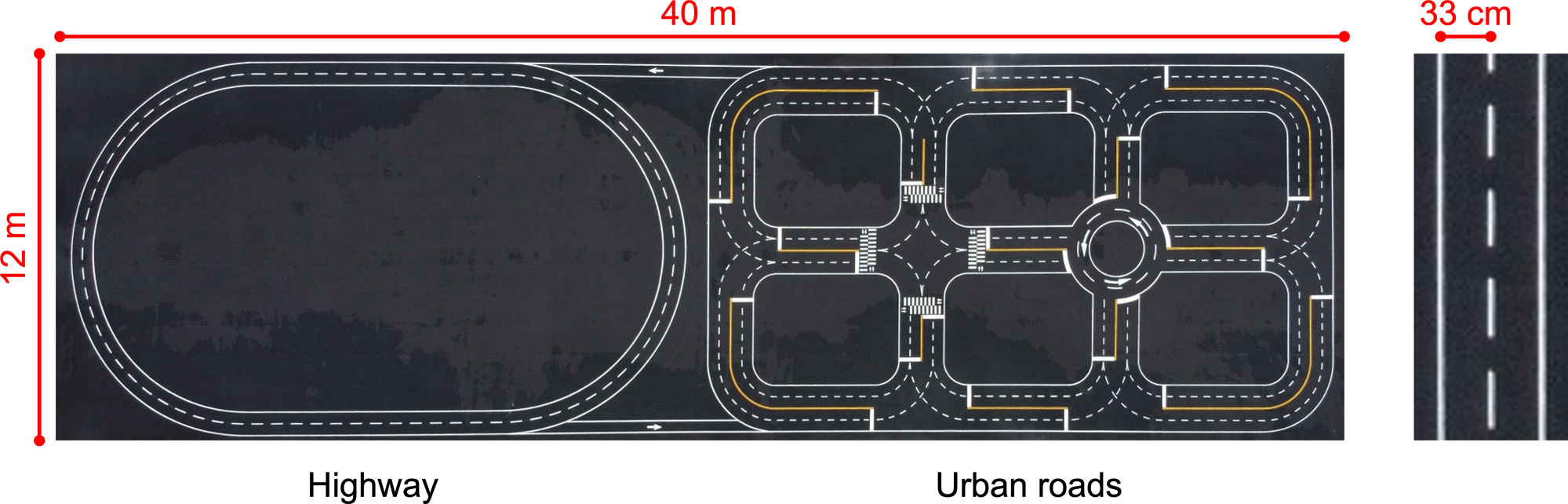}}
\caption{Scale truck platooning proving ground in Cyclops.}
\label{fig:pg}
\end{figure}

With this motivation, we developed a scale truck platooning testbed, as shown in Fig.~\ref{fig:testbed}, which is named {\em cyclops} since the single front-facing camera on the tractor unit reminds us of the one-eyed giant in Greek mythology. Our testbed includes three autonomous semi-trailer trucks, each of which is equipped with perception sensors, computing platforms, and wireless communication devices. With the truck platforms, we developed a platooning system with our perception and control systems that closely resemble the full-size platooning system. It retains most of the real-world challenges in platooning but significantly reduces the cost of implementing safety scenarios. Thus, it allows us to test and validate safety countermeasures with lower cost in the early development stage, even before the full-size trucks are ready. The platooning system can operate in our scale truck proving ground, as shown in Fig.~\ref{fig:pg}, where various platooning scenarios can be rapidly developed and validated.


\begin{figure*}
\vspace*{0.2 cm}
\centerline{\includegraphics[scale=0.5]{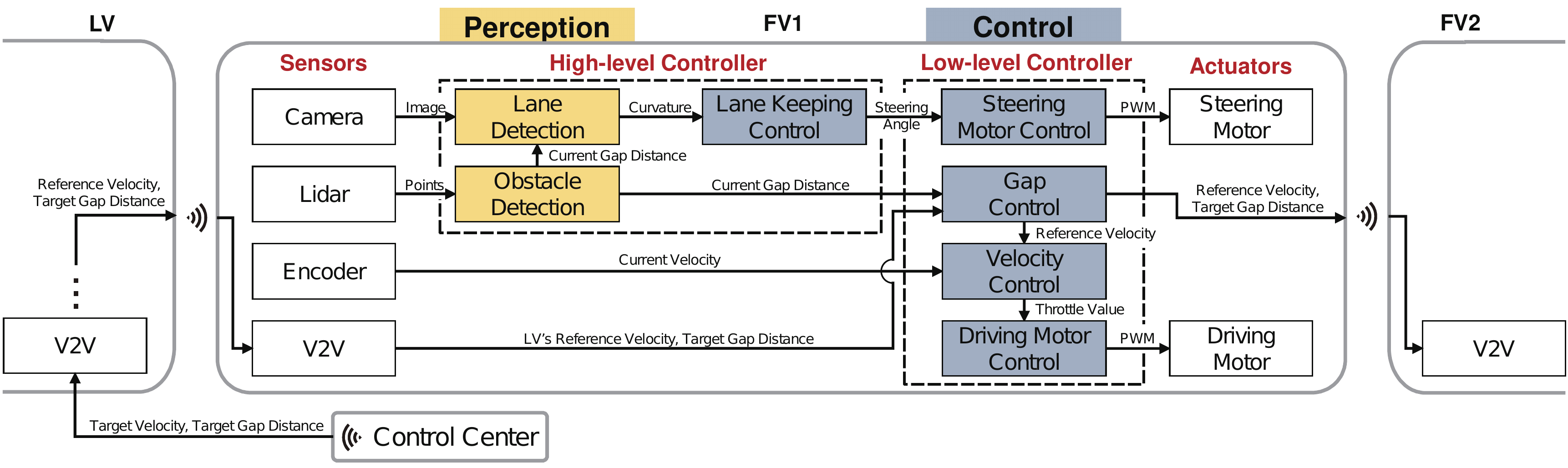}}
\vspace{-0.3cm}
\caption{Cyclops architecture for platooning.}
\label{fig:system}
\end{figure*}

The perception system employs one camera sensor and one single-plane lidar sensor. The camera detects ego lanes for lateral control, while the lidar measures the longitudinal distance and the lateral location of a preceding trailer. Regarding the perception system, we were attracted by a unique challenge of truck platooning. Since FVs closely follow their preceding trailers, the rear end of trailers significantly occludes the camera's view, as was also reported in the case of full-size trucks~\cite{kim2020camera}. The occlusion adversely affects the lane detection performance because the trailer's rear end creates indistinguishable noises. To reduce such noises, our perception system dynamically adjusts the region of interest (ROI) in the camera image based on the distance to the preceding trailer, keeping the ROI free from the noises.

For the platooning control, the following three algorithms are developed: (i) velocity control, (ii) gap control, and (iii) lane keeping control. The first is to have the vehicle move at the reference velocity. The second is to maintain the distance from the preceding truck to the desired gap reference. Gap control utilizes the distance to the preceding trailer from the lidar sensor and the reference velocity of the preceding truck through wireless communications. Finally, the third utilizes the camera sensor to locate the ego truck's lateral location, adjusting the steering angle.

We conducted a case study with a camera sensor failure with the developed testbed. Since our platooning system relies on the camera for lateral control, a camera failure leads to a catastrophic event. Considering this hazardous safety scenario, we developed a mitigation method that temporarily utilizes the lidar sensor such that FV follows the center point of the preceding trailer instead of following the lane until a graceful stop. With this case study, we argue that our testbed is useful in validating safety algorithms and platooning scenarios.

We share in this paper our experience gained while developing the scale truck platooning testbed and release all the implementation details, including the algorithm source codes\footnote{All the design materials and the source codes are available at \url{https://github.com/hyeongyu-lee/scale_truck_control}.}. We hope that our testbed can be reproduced by other research groups interested in truck platooning systems.

\section{Truck and System Design}
\label{sec:design}

\subsection{System Design}
\label{sec:system}

Fig.~\ref{fig:system} shows the overall architecture of our testbed, where three trucks (LV, FV1, and FV2) are connected through a wireless network to each other and also with a remote control center. The control center communicates directly with LV, providing the target velocity and gap distance. The target velocity is used as the reference velocity of LV, while the target gap distance is communicated to FV1, which is relayed to FV2 for velocity and gap control. We point out that every truck in Cyclops is autonomous, which is not valid in full-size truck platooning, where its LV is usually human-driven.

In FV1 and FV2, the perception system receives the front-facing camera images and the lidar data, where the camera image is used for detecting the lane curvature, while the lidar points are used to locate the preceding trailer's rear end. Then the following information is delivered to the control system: (i) the lane curvature, (ii) the gap distance, (iii) the current velocity from the encoder sensor, (iv) the target gap distance given by the command center, and (v) the preceding truck's reference velocity. Now the lane keeping controller decides the steering angle, controlled by the steering motor through the steering motor controller. The gap controller generates the reference velocity to meet the target gap, then given to the velocity controller that controls the driving motor through the driving motor controller. The reference velocity generated by the gap controller is communicated to the following truck through V2V communication for its gap control.


LV and FVs share the same hardware and software architecture with a slight difference. Since LV does not need the gap controller, it is replaced with the emergency braking controller between the obstacle detection and the velocity controller. Since our testbed does not allow any other surrounding vehicles except the trucks at this moment, LV automatically enforces an emergency stop when it detects an obstacle in the driving direction, which in turn triggers emergency stops in FV1 and FV2. Additionally, the command center can also enforce an emergency stop by a remote command.

\begin{table}[t] 
\vspace{0.4cm}
\caption{Summary of Truck Components. \label{tab:components}}
\begin{center}
\vspace{-0.3cm}
\begin{tabular}{l|l}
\hline \hline
{\bf Component}         & {\bf Item} \\
\hline
Tractor                 & 1/14 MAN 26.540 6x4 XLX \\
Trailer                 & 40-Foot Container 1/14 Semi-trailer \\
Camera                  & ELP-USBFHD04H-BL180 \\
Lidar                   & RPLidar A3 \\
Encoder                 & Magnetic Encoder Pair Kit \\
High-level controller   & Nvidia Jetson AGX Xavier \\
Low-level controller    & OpenCR 1.0 \\
Driving Motor           & TBLM-02S 21.5T \\
Steering motor          & SANWA SRM-102 \\
Battery                 & BiXPower 96Wh BP90-CS10 \\
Antenna                 & ipTIME N007 \\
\hline \hline
\end{tabular}
\end{center}
\end{table}

\begin{figure}
\centerline{\includegraphics[scale=0.27]{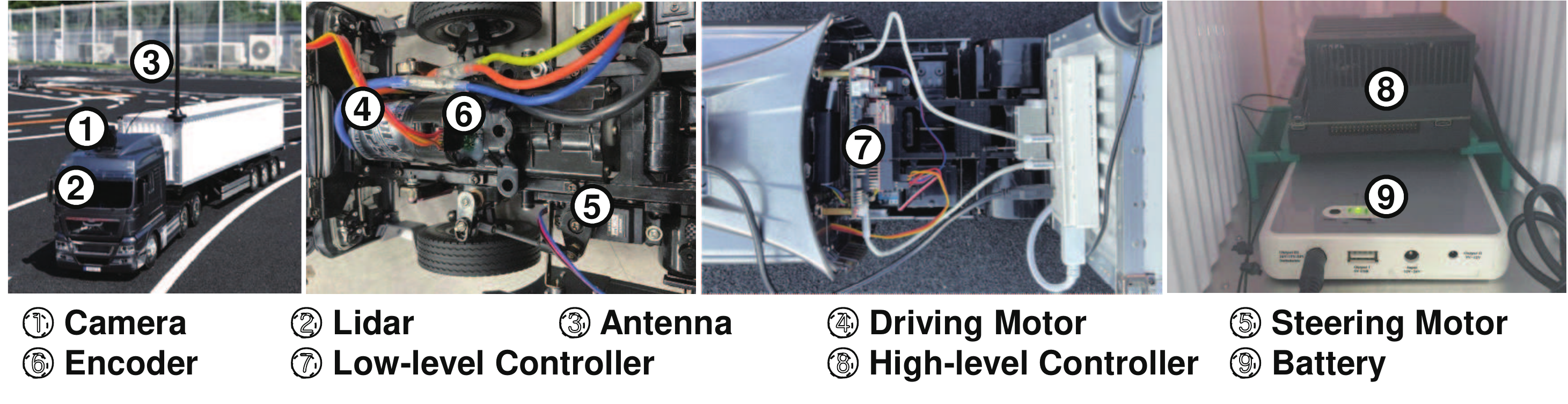}
\vspace{-0.3cm}}
\caption{Scale truck design.}
\label{fig:truck}
\end{figure}
\vspace{0.3cm}

\subsection{Truck Design}
\label{sec:truck}

Fig.~\ref{fig:truck} shows one of our 1/14 scale semi-trailer trucks. As the truck body, we use an RC truck famous among the RC enthusiasts. Three such trucks were developed, one designated as LV while the others are FVs. The figure highlights nine essential components such as sensors and computing platforms. The hardware components are listed in Table~\ref{tab:components}. More specifically, each truck is equipped with a 180 degrees FoV (Field of View) front-facing camera on top of the tractor. A single-plane lidar sensor with a 25~m detection range is hidden inside the truck cabin. An encoder that counts wheel rotations is installed at the tractor's front wheel to measure the vehicle velocity. We use two heterogeneous computing platforms for the high-level and low-level control, respectively. For the high-level controller, we use an Nvidia Jetson AGX Xavier located inside the trailer, while, as the low-level controller, an OpenCR embedded computer is attached at the back of the tractor. For the driving motor, we use a 1800~KV BLDC motor. For the steering motor, we use a servo motor with a torque of 3.3~kg$\cdot$cm. All the electronic components are powered by a 96~Wh battery in the trailer. We use the WIFI network interface in the Nvidia Jetson AGX Xavier for wireless communication, which is reinforced by an additional high-performance antenna installed on the trailer.

\section{Perception and Control}
\label{sec:perception}

This section presents our perception and control methods for truck platooning. Our platooning system has two different modes of operations (i.e., the LV mode and the FV mode). We implicitly assume the FV mode in this section, and it will be explicitly stated when the LV mode needs to be presented.

\subsection{Lane Detection}
\label{sec:lane}

\begin{figure}
\vspace{0.2cm}
    \centering
    \subfigure[0.8~m gap with static ROI.]{
    \includegraphics[width=6.5cm]{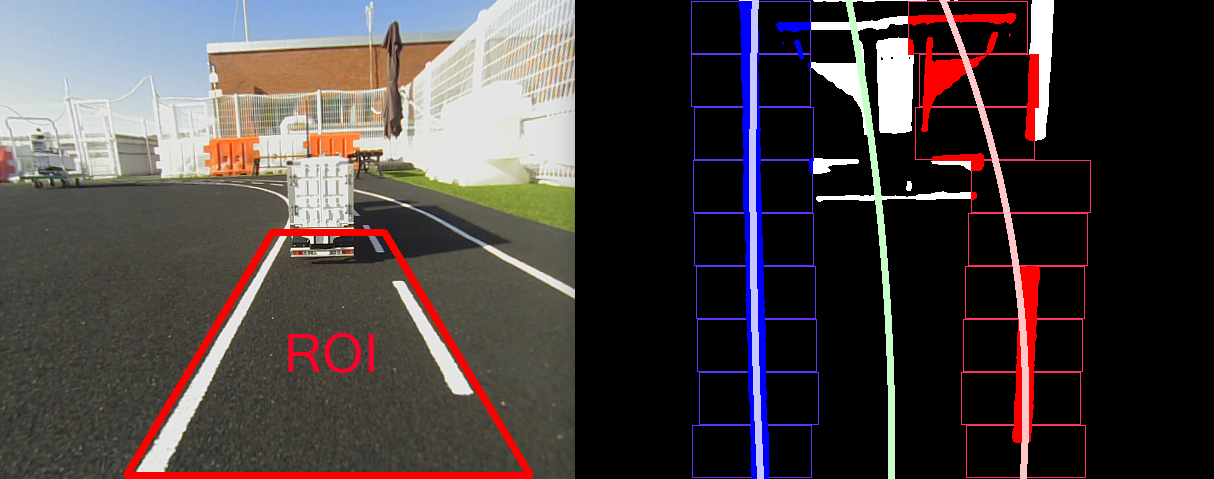}}
    \subfigure[0.8~m gap with dynamic ROI.]{
    \includegraphics[width=6.5cm]{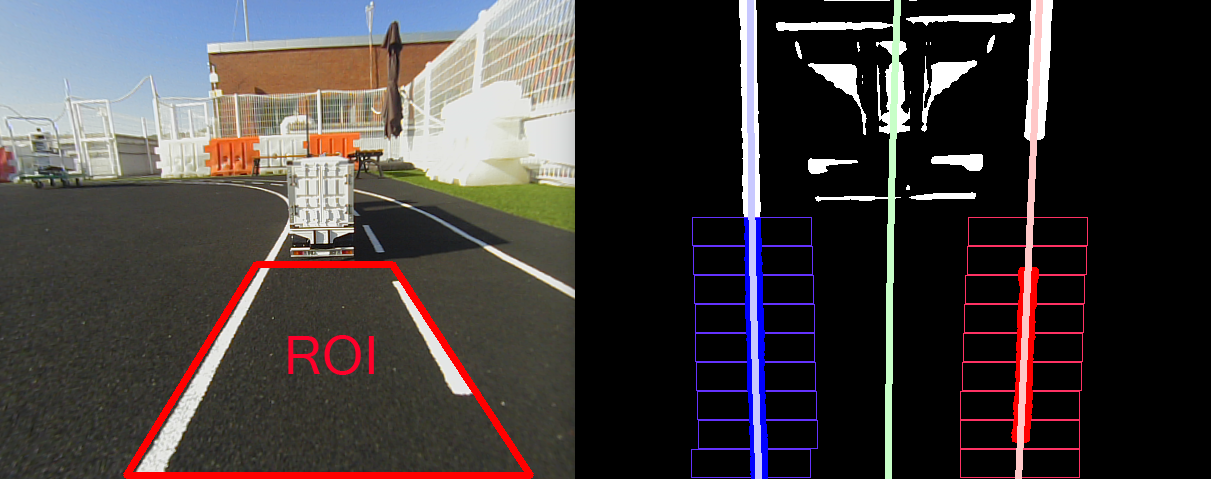}}
    
\vspace{-0.2cm}
\caption{Lane detection performance with/without dynamic ROI in a small (0.8~m) inter-vehicle gap situation.}
\label{fig:roi}
\vspace{-0.2cm}
\end{figure}

The objective of lane detection is to find the curvature of the desired path from the front-facing camera. For that, a camera image goes through the following steps: (i) pre-processing of cutting the trapezoid ROI and removing noises using the Gaussian blur method; (ii) conversion into a binarized bird's eye view image; (iii) sliding window-based search to detect the left and right lanes; (iv) fitting the desired path's curvature combining the left and right lane curvatures. The final curvature is decided as a second-order polynomial.

The above lane detection method has been widely used in many studies~\cite{fan2011robust, zhang2018combined, haque2019computer} because it is robust and lightweight for the implementation in small embedded systems. Unfortunately, however, the method does not perform well in the platooning scenario due to the minimal gap to the preceding truck. Fig.~\ref{fig:roi}(a) shows a forward view image and its lane detection result with the static ROI method where the trapezoid ROI is a fixed one. The binary image indicates that the slight inclusion of the trailer's rear end in the ROI badly affects the lane detection result. The noise by the preceding trailer is a known issue reported in the platooning with full-size trucks~\cite{kim2020camera}. In our testbed, the effect is even vibrant because the trailer's color happens to be white.

To eliminate the noise by the trailer at a close distance, we try to cut off the trailer portion in the bird's eye view image. For that, we use the gap information from the lidar sensor. By actual measurements, we found that the ratio between the longitudinal distance and the number of vertical pixels is 490 pixels per 1.0~m. With this mapping, Fig.~\ref{fig:roi}(b) shows the adapted ROI that precisely removes the noise, which provides a significantly improved lane detection result. This adaptation logic is added in our lane detection module such that it can dynamically determine the optimal ROI.

\subsection{Obstacle Detection}
\label{sec:target}

Obstacle detection provides the gap distance to the preceding truck. We use a single-plane lidar sensor that provides a set of points to the nearest objects around 360 degrees. To detect the preceding truck, we limitedly use the front-facing 60 degrees. For the obstacle detection, we employ the well-known obstacle\_detector ROS package~\cite{obstacle_detector}, which detects and also tracks nearby objects by clustering the 2D lidar points. The tracking result is obtained as a circle where its center and diameter correspond to the middle point and the horizontal length of the preceding trailer's rear end, respectively.

\subsection{Velocity Control}
In this section, we introduce the velocity control, which consists of the feed-forward plus anti-windup proportional-integral (FF+AWPI) control, the velocity saturation function ${\rm sat}_{i}(u_{v,c,i})$, and the inverse of the nonlinear function $f_{i}^{-1}(\bar{u}_{v,c,i})$. Here, the subscript $i = 0$ denotes the variables or function for LV, $i=1$ for FV1, and $i=2$ for FV2. The FF+AWPI control calculates the velocity control input $u_{v,c,i}$ such that
\begin{equation} \label{eq:u_vci}
\begin{split}
u_{v,c,i}(k)&= K_{F,i}v_{r,i}(k)+K_{P,i}\tilde{v}_{i}(k)+\sum_{j=0}^{j=k}{K_{I,i}\tilde{v}_{i}(j)T_{s}}\\
& \ \ \ + \sum_{j=0}^{j=k-1}{K_{A,i}\left(\bar{u}_{v,c,i}(j)-u_{v,c,i}(j)\right)},
\end{split}
\end{equation}
where $T_{s}$ is the sampling time, $\bar{u}_{v,c,i}$ is the velocity control input saturated by function ${\rm sat}_{i}(u_{v,c,i})$, the positive constants $K_{P,i}, K_{F,i},K_{I,i}$, and $K_{A,i}$ are control parameters, and $\bar{v}_{r,i}$ is the saturated velocity reference. The velocity tracking error $\tilde{v}_{i}$ is given as
\begin{equation} \label{eq:tilde_vi}
\begin{split}
\tilde{v}_{i}(k)=\bar{v}_{r,i}(k)-v_{i}(k),
\end{split}
\end{equation}
where $v_{i}$ is the velocity measurements of the truck.

The saturation function ${\rm sat}_{i}(u_{v,c,i})$ represents the velocity limitation that a truck can generate. The saturation function ${\rm sat}_{i}(u_{v,c,i})$ is given as
\begin{equation} \label{eq:sat} 
\begin{split} 
{\rm sat}_{i}(u_{v,c,i}) = 
\begin{cases} 
V_{i}^{\max}, & {\rm if} \ u_{v,c,i} \geq V_{i}^{\max} 
\\ 0 , & {\rm if} \ u_{v,c,i} \leq 0
\\ u_{v,c,i}, & {\rm otherwise}
\end{cases},
\end{split}
\end{equation}
where $V_{i}^{\max}$ denotes the maximum velocity that a truck can generate.

The motor command and the velocity of the truck have the nonlinear relation. The inverse of nonlinear function $f_{i}^{-1}(\bar{u}_{v,c,i})$ is used for linearly treating the truck. The nonlinear function $f_{i}(u_{i})$ representing the nonlinearity between the motor command and the velocity of the truck is obtained by fitting the experimental data for the motor command and the velocity of truck into a quadratic function, and is given by
\begin{equation} \label{eq:nonlinear_function} 
\begin{split} 
f_{i}(u_{i}) = a_{i}u_{i}^{2} + b_{i}u_{i} + c_{i},
\end{split}
\end{equation}
where $a_{i}, b_{i}$, and $c_{i}$ are design parameters. From \eqref{eq:nonlinear_function}, we can induce the function $f_{i}^{-1}(\bar{u}_{v,c,i})$ such that
\begin{equation} \label{eq:inverse_nonlinear_function} 
\begin{split} 
f_{i}^{-1}(\bar{u}_{v,c,i}) = \frac{-b_{i}+\sqrt{b_{i}^{2}-4a_{i}(c_{i}-\bar{u}_{v,c,i})}}{2a_{i}}.
\end{split}
\end{equation}
Finally, the output of the function $f_{i}^{-1}(\bar{u}_{v,c,i})$ is the motor command $u_{i}$, which is applied to the driving motor.

\subsection{Gap Control}
The gap control is used by only FVs. The goal of the gap control is to make the gap to the preceding truck closely follow the gap reference. The gap control is composed of the feed-forward plus proportional-differential (FF+PD) control and the velocity reference saturation function ${\rm sat}_{r,i}(v_{r,i})$. The FF+PD control is given in the following form:
\begin{equation} \label{eq:gap control} 
\begin{split} 
v_{r,i}(k) &= \bar{v}_{r,i-1}(k) - K_{GP,i}\tilde{d}_{i}(k) \\ 
& \ \ \ -\frac{K_{GD,i}(\tilde{d}_{i}(k)-\tilde{d}_{i}(k-1))}{T_{s}},
\end{split}
\end{equation}
where $\bar{v}_{r,i-1}$ is the saturated velocity reference of the preceding truck received via wireless communication. The positive constants $K_{GP,i}$ and $K_{GD,i}$ are control parameters, and the gap tracking error $\tilde{d}_{i}$ is obtained as
\begin{equation} \label{eq:gap gracking error} 
\begin{split} 
\tilde{d}_{i}(k) = d_{r,i}(k)-d_{i}(k),
\end{split}
\end{equation}
where $d_{r,i}$ denotes the gap reference and $d_{i}$ denotes the measured gap to the preceding truck. 

The velocity reference saturation function ${\rm sat}_{r,i}(v_{r,i})$ saturates the velocity reference as the velocity limit of the lane, which is represented as
\begin{equation} \label{eq:sat_reference} 
\begin{split} 
{\rm sat}_{i}(v_{r,i}) = 
\begin{cases} 
V_{r}^{\max}, & {\rm if} \ v_{r,i} \geq V_{r}^{\max} 
\\ 0 , & {\rm if} \ v_{r,i} \leq 0
\\ v_{r,i}, & {\rm otherwise}
\end{cases},
\end{split}
\end{equation}
where $V_{r}^{\max}$ denotes the velocity limit of the lane. Finally, the saturated velocity reference $\bar{v}_{r,i}$ plays as the velocity reference for the velocity control.

\subsection{Lane Keeping Control}
In this section, we introduce the lane keeping control to maintain the truck within the lane. The lane keeping control is given by
\begin{equation} \label{eq:lane_keeping_control} 
\begin{split} 
\delta_{i}(k) = -K_{i}(v_{i})e_{i}(k) - K_{L,i}(v_{i})e_{L,i}(k),
\end{split}
\end{equation}
where $\delta_{i}$ is the steering angle, $e_{i}$ is the estimated lateral error, $e_{L,i}$ is the preview lateral error, and $K_{i}(v_{i})$ and $K_{L,i}(v_{i})$ are control parameters that are varied according to the truck velocity. Because faster velocities lead to bigger lateral changes by a given steering angle, with a faster velocity, the control parameters have to be smaller.

In \eqref{eq:lane_keeping_control}, the preview lateral error $e_{L,i}$ is obtained with a camera sensor. The lateral error $e_{i}$, however, cannot be directly obtained with the camera sensor due to its limited sight, thus we estimate the lateral error $e_{i}$ such that
\begin{equation} \label{eq:lateral_position_error} 
\begin{split} 
e_{i}(k) = e_{L,i}(k) - L_{i}{\rm tan}(\theta_{i}(k)),
\end{split}
\end{equation}
where $L_{i}$ is the preview distance and $\theta_{i}$ is the angle between the direction in which the truck moves and the preview point on the desired path which is obtained with the camera sensor.

\section{Experiments}\label{sec:experiments}
This section first evaluates the platooning performance of our developed scale truck platforms with two scenarios. The first scenario is to speed up LV while maintaining the constant gap between trucks, and the second one is to reduce the gap between trucks while maintaining the constant velocity of the platoon. Then we evaluate the effectiveness of our dynamic ROI method by measuring how it affects lateral control stability. However, as mentioned in the previous section, the front-facing camera sensor cannot effectively measure the lateral error, which is essential for evaluating the lane keeping control. As an alternative, we install an extra camera sensor, downfacing each trailer's side, which is dedicated to measuring the lateral error.

\begin{figure}[t]
\vspace{-0.5cm}
\centering
\includegraphics[width=3.4in]{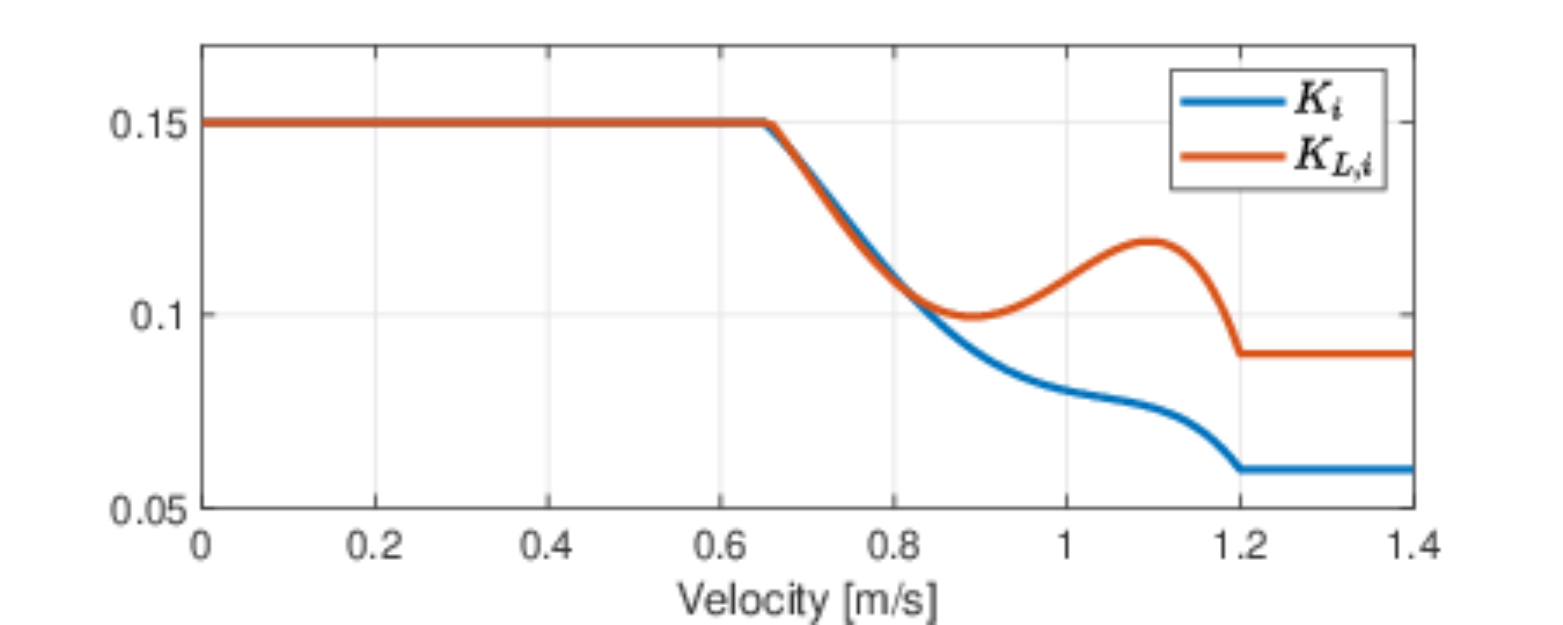}
\caption{Lane keeping control parameters in the experiments. \label{fig:LKC_parameters}}
\vfill
\end{figure}

For all the trucks, the velocity control parameters in \eqref{eq:u_vci} are identically chosen as $K_{F,i} = 1, \ K_{P,i} = 0.8, \ K_{I,i} = 2.0$, and $ K_{A,i} = 0.0001$ for $i = 0,\ 1,\ 2$. The parameters $a_{i}, b_{i}$ and $c_{i}$ in \eqref{eq:nonlinear_function} are selected as Table~\ref{table:nonlinear_function_parameters}. The gap control parameters of FV1 and FV2 in \eqref{eq:gap control} are also identically chosen as $K_{GP,i} = 0.5$ and $K_{GD,i} = 0.1$ for $i = 1, \ 2$. The lane keeping control parameters of all the trucks are designed as shown in Fig.~\ref{fig:LKC_parameters}. The parameter $V_{i}^{\max}$ for $i = 0, \ 1, \ 2$ in \eqref{eq:sat} is selected as 2~m/s, and $V^{\max}_{r}$ in \eqref{eq:sat_reference} is chosen as 1.4~m/s. In the following graphs, the red colored areas commonly indicate that LV is in the curved road segments, whereas the white areas indicate straight road segments.

\begin{table}[t] 
\vspace{0.4cm}
\caption{Parameters for Nonlinear Function.\label{table:nonlinear_function_parameters}}
\begin{center}
\vspace{-0.3cm}
\begin{tabular}{c|c|c|c}
\hline \hline
\begin{tabular}{p{0.1in}} $i$ \end{tabular}&\begin{tabular}{p{0.1in}} $a_{i}$
\end{tabular}&\begin{tabular}{p{0.1in}} $b_{i}$ \end{tabular} & \begin{tabular}{p{0.1in}} 
$c_{i}$ \end{tabular}\\ \hline
$0$ & $-1.1446 \times 10^{-5}$ & $0.048278$ & $-47.94$ \\
$1$ & $-2.0975 \times 10^{-5}$ & $0.08152$ & $-76.87$ \\
$2$ & $-1.0444 \times 10^{-5}$ & $0.043253$ & $-42.3682$ \\
\hline \hline
\end{tabular}
\end{center}
\end{table}

\begin{figure}[!t]
\centering
\vspace{0.2cm}
\subfigure[Velocity control performance.]
{\includegraphics[width=3.4in]{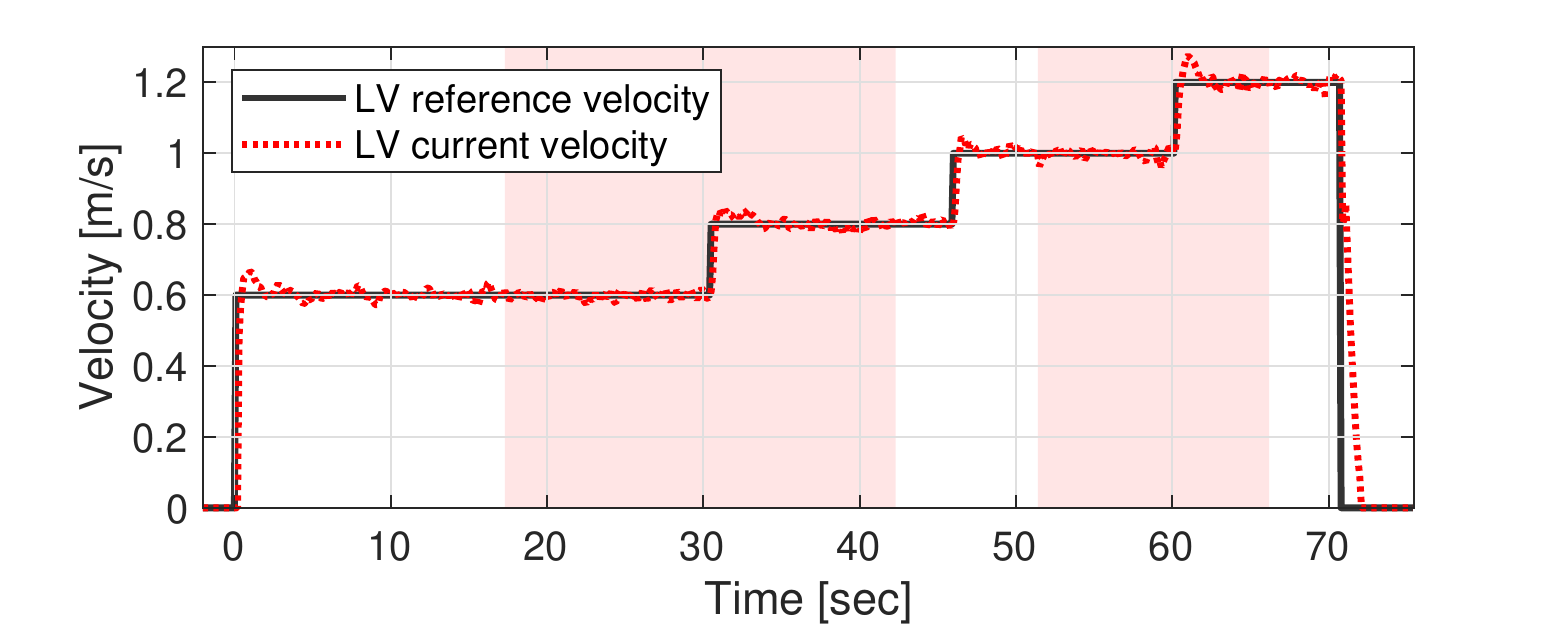} \label{fig:Scenario1_vel}}

\vspace{-0.3cm}
\subfigure[Gap control performance.]
{\includegraphics[width=3.4in]{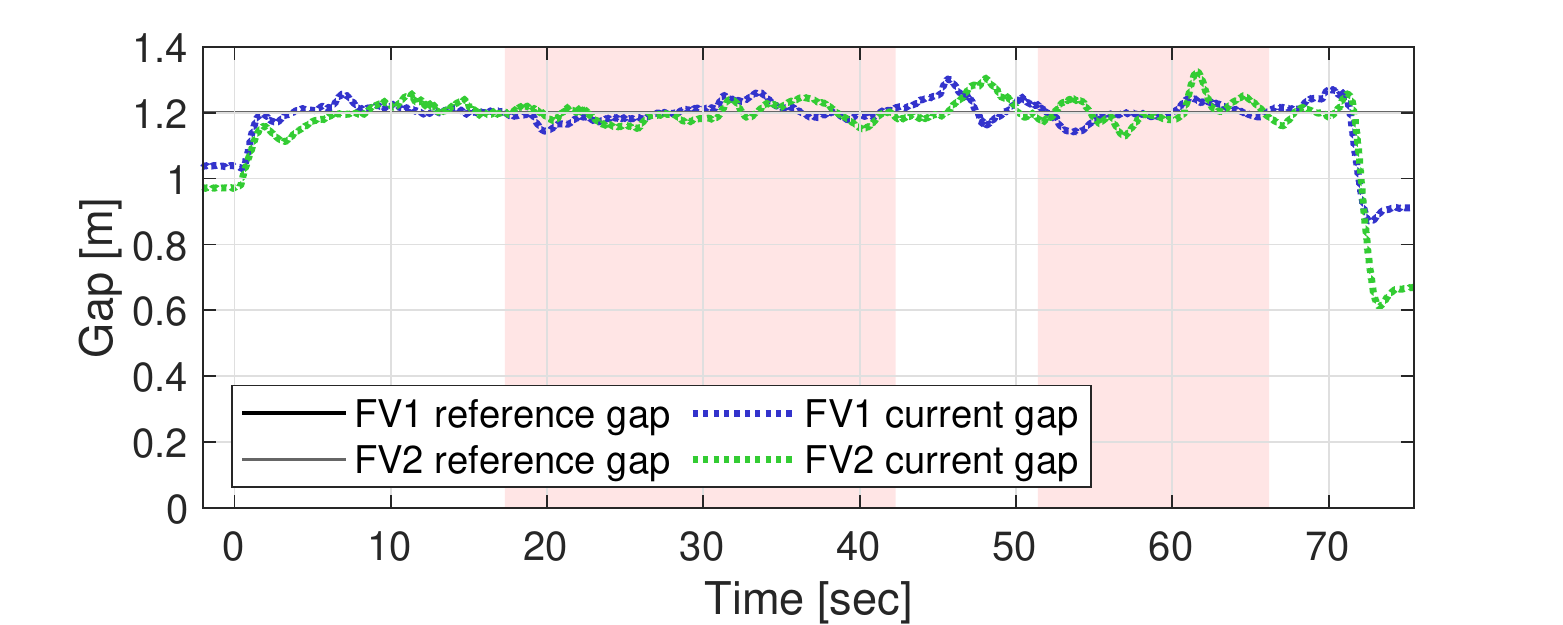} \label{fig:Scenario1_gap}}

\vspace{-0.3cm}
\subfigure[Lane keeping control performance.]
{\includegraphics[width=3.4in]{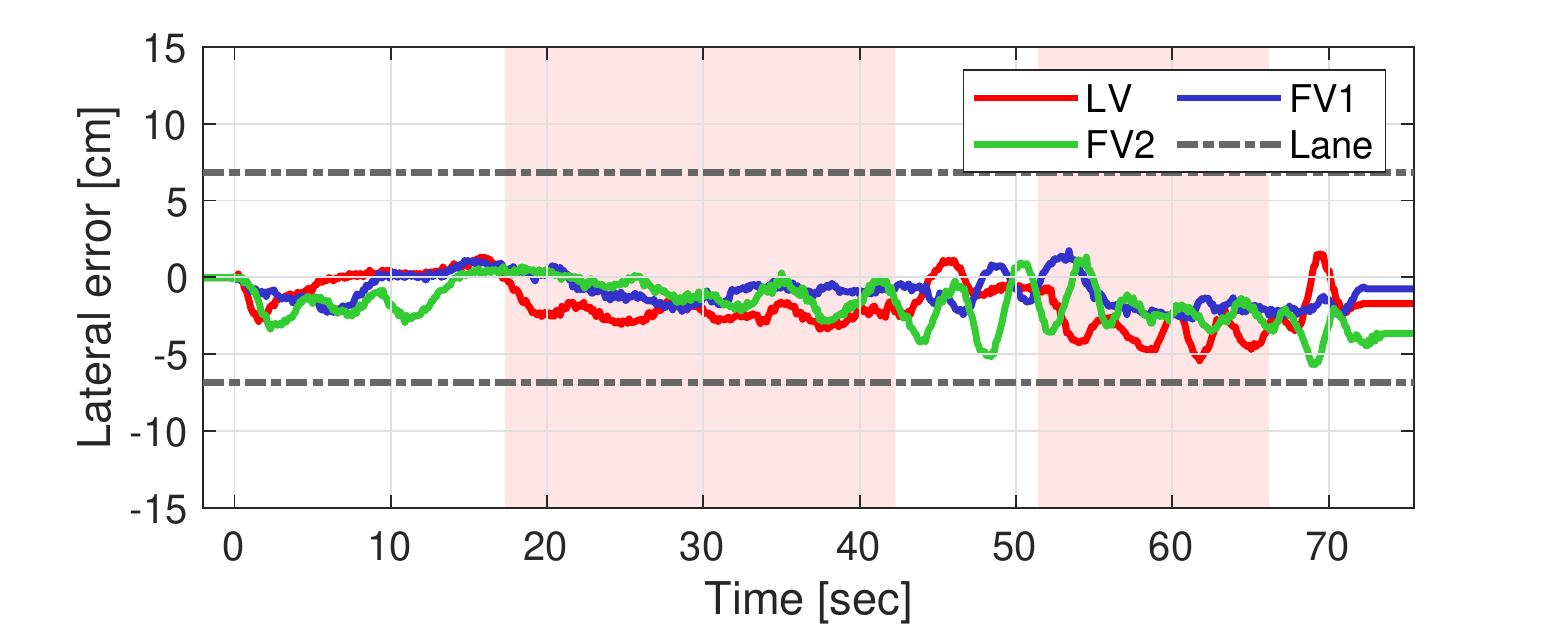} \label{fig:Scenario1_lateral_error}}
\caption{Experimental result platooning performance in Scenario 1.}
\label{fig:Scenario1}
\end{figure}

\begin{figure}[!t]
\centering
\vspace{0.2cm}
\subfigure[Velocity control performance.]
{\includegraphics[width=3.4in]{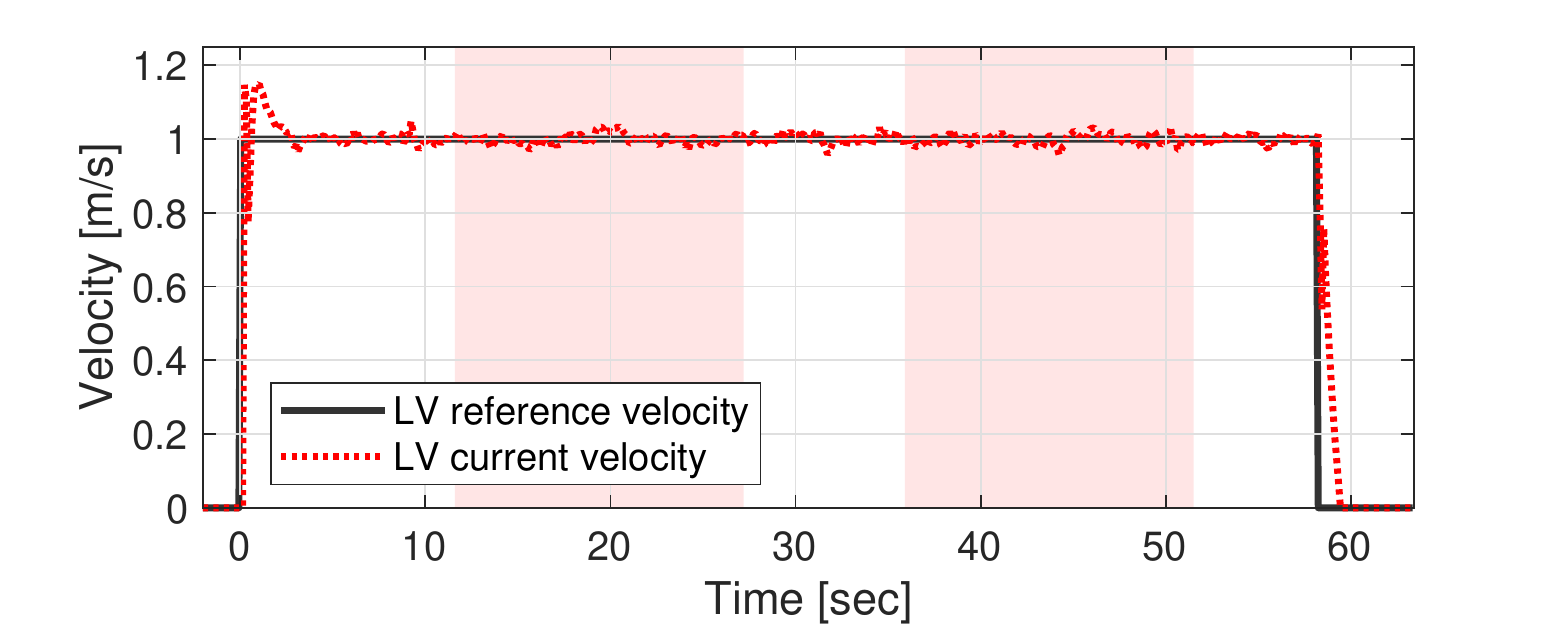} \label{fig:Scenario2_vel}}

\vspace{-0.3cm}
\subfigure[Gap control performance.]
{\includegraphics[width=3.4in]{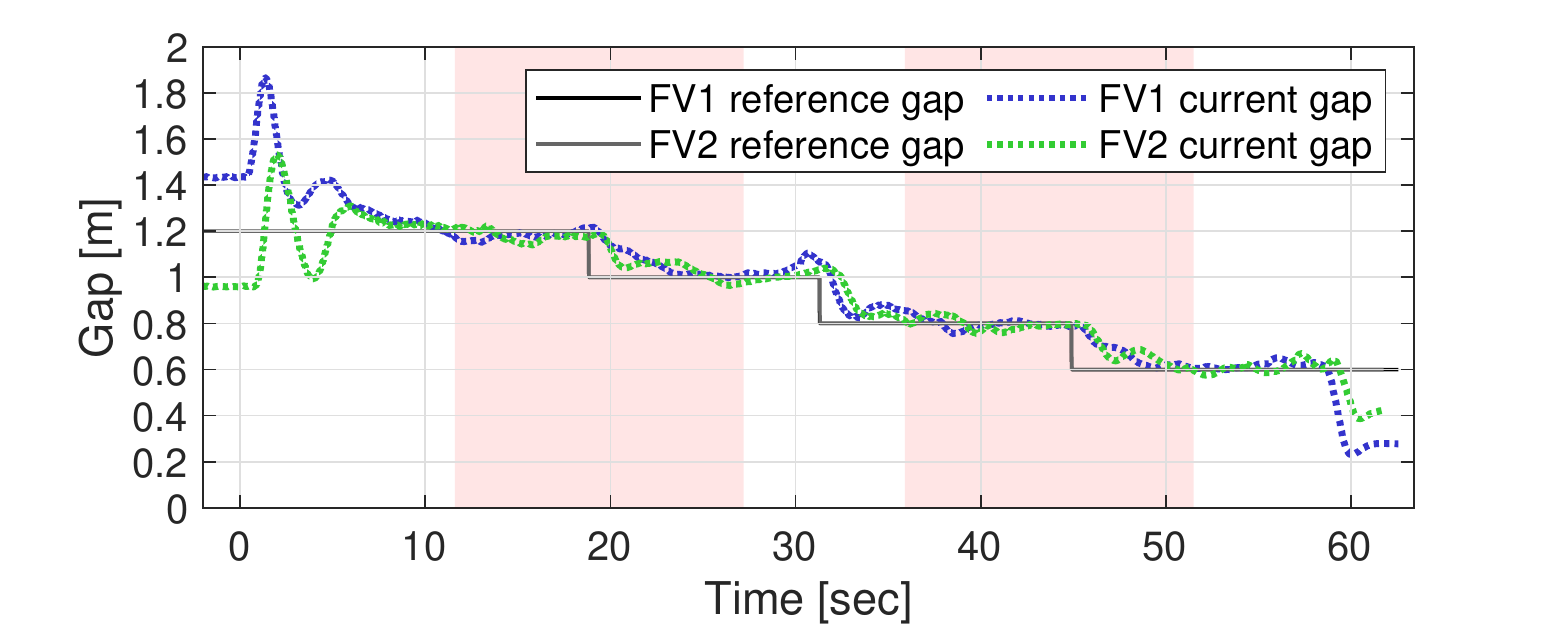} \label{fig:Scenario2_gap}}

\vspace{-0.3cm}
\subfigure[Lane keeping control performance.]
{\includegraphics[width=3.4in]{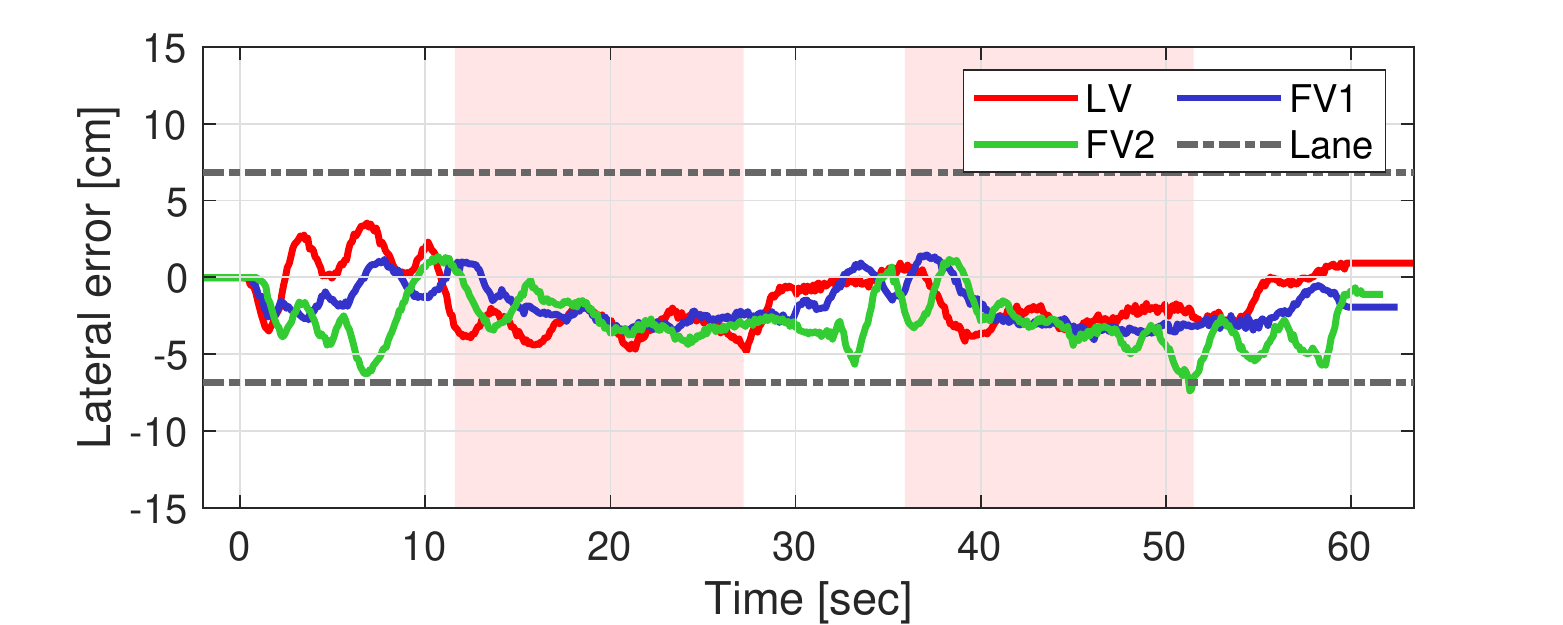} \label{fig:Scenario2_lateral_error}}
\caption{Experimental result platooning performance in Scenario 2.}
\label{fig:Scenario2}
\end{figure}

\subsection{Performance of Platooning with Three Trucks}

Fig.~\ref{fig:Scenario1} shows the evaluation results for the first platooning scenario (Scenario 1). Fig.~\ref{fig:Scenario1_vel} shows how LV follows the given reference velocity as it ramps up from 0.6~m/s to 1.2~m/s. Although the results show slight overshoots upon velocity changes, they do not last long, resulting in satisfactory longitudinal control performance. Fig.~\ref{fig:Scenario1_gap} shows the measured gaps between trucks as the reference gap is constant at 1.2~m. The results show that all the trucks maintain the desired gap well despite the acceleration of the platoon. The lateral errors, meanwhile, increase upon the velocity goes about 1~m/s, but all the trucks are safely maintained within the lane.

In Fig.~\ref{fig:Scenario2}, the second scenario (Scenario 2) is validated, where the reference gap ramps down from 1.2~m to 0.6~m while maintaining a constant velocity at 1.0~m/s. Fig.~\ref{fig:Scenario2_vel} shows the LV velocity well maintained at 1.0~m/s. Fig.~\ref{fig:Scenario2_gap} shows that the measured gaps stick well to the dynamic reference gap. However, in Fig.~\ref{fig:Scenario2_lateral_error}, we can see that the lateral errors get significant while the reference gap is the smallest (i.e., 0.6~m). As mentioned before, this lateral safety problem happens when the gap is too short such that the camera's view is significantly occluded by the preceding trailer. 

\subsection{Effectiveness of Dynamic ROI}

\begin{figure}[!t]
\centering
\includegraphics[width=3.4in]{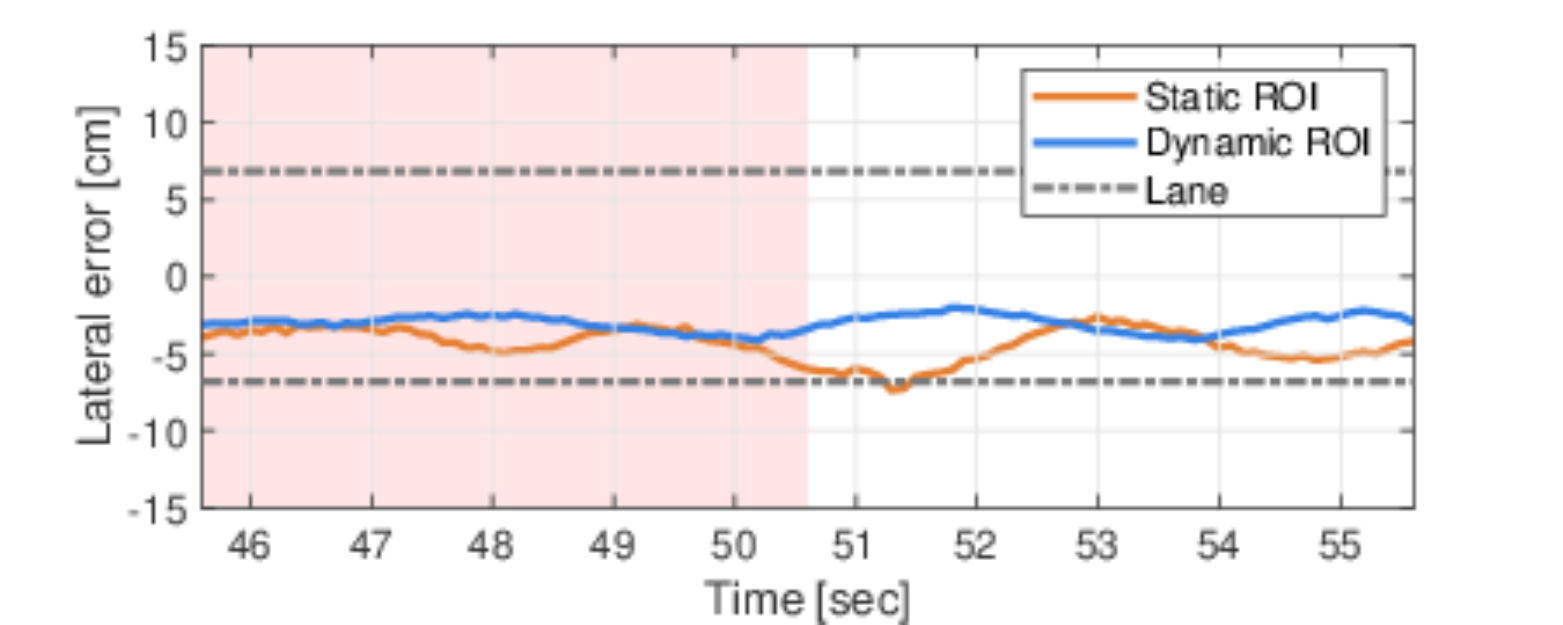}
\caption{Lane keeping performance with/without using dynamic ROI in camera image processing.}
\label{fig:DynamicROI}
\end{figure}

We apply the dynamic ROI method to the lane detection module for solving the lateral stability problem. Fig.~\ref{fig:DynamicROI} compares FV2's lateral errors with static ROI and dynamic ROI while maintaining the 0.6~m gap between FV1 and FV2. In the figure, the lateral error with dynamic ROI is visibly smaller than that with static ROI. Even worse, FV2 goes out the lane at the 51.4-second point when operated with the static ROI method. In contrast, the truck is safely maintained within the lane with dynamic ROI. Thus, we claim that dynamic ROI is an effective method to enhance lateral stability and safety.

\section{Case Study: Camera Sensor Failure}
\label{sec:case}
To show how our testbed can help develop safety scenarios, we conducted a case study of developing a mitigation method for a camera sensor failure scenario on our testbed. Since the lateral control depends on the camera sensor, its failure can be catastrophic. In such a fatal scenario, one possible method is to use the remaining lidar sensor to follow the preceding trailer's center, even without knowing the lane curvature. By this method, the platoon can buy enough time until it slowly stops while maintaining minimal safety.

For the experiment, we implemented a fault injection scenario that simulates a cut wire from the camera in FV1. In such cases, the old image is repeatedly fed into the perception system such that the failure can be detected by comparing the current image with old images assuming that the truck's velocity is not zero. Then the truck goes into a fail-safe mode that no longer uses the lane detection result but follows the preceding truck's center. The failure is also notified to LV such that the entire platoon goes into a fail-safe mode, slowing down until a graceful stop.

\begin{figure}[!t]
\vspace{0.2cm}
\centering
\includegraphics[width=3.4in]{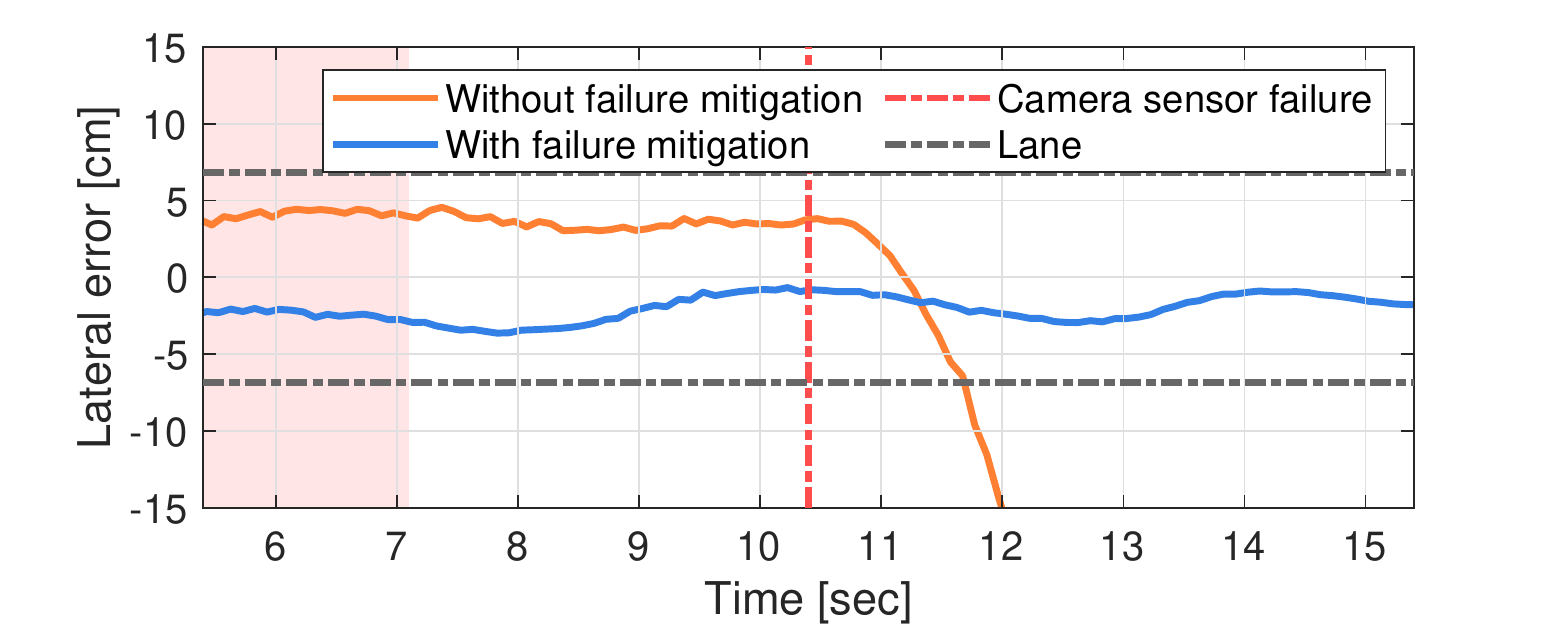}
\caption{Lane keeping performance with/without the camera failure mitigation method.}
\label{fig:CameraFailure}
\end{figure}

Fig.~\ref{fig:CameraFailure} compares FV1's lateral error for two scenarios. One is without any failure mitigation method, and the other is with our failure mitigation method. For the measurement, the platoon drives at a velocity of 0.8~m/s with a gap of 0.8~m. After LV reaches a straight road segment, a fault is injected into FV1's camera. The figure shows that the truck quickly goes out of the lane without the mitigation method, whereas our mitigation method makes it keep its lane even though the lateral oscillation is slightly amplified.

\section{Related World}
\label{sec:related}
The truck platooning technology has been developed through national flagship projects~\cite{bergenhem2012overview, tsugawa2016review, robinson2010operating, bergenhem2010challenges, tsugawa2011automated, michael1998capacity, carbaugh1998safety}. Most studies thus far have dealt with control issues~\cite{rajamani2000demonstration, ploeg2013controller, alam2015experimental, besselink2017string}, fuel economy~\cite{alam2015heavy, turri2016cooperative, al2010experimental}, and safety issues~\cite{van2016towards, bijlsma2017fail, van2017evaluation}. To the best of our knowledge, our study is one of the first attempts to implement a scale truck platooning tested.


\section{Conclusion}
\label{sec:conclusion}

We presented Cyclops, a scale truck platooning testbed, targeting an open research platform for everyone interested in the truck platooning technology. Our platooning system uses wireless communications as well as camera and lidar sensors to maintain a constant gap at a given platooning velocity. We implemented our perception and control systems optimized for solving the platooning system's unique challenges caused by the minimal gap between trucks. With our scale trucks, we successfully demonstrated the control stability up to 1.2~m/s, corresponding to about 60~km/h in actual trucks. Also, we can safely reduce the gap until it reaches 0.6~m, which corresponds to 8.4~m in actual trucks, at the velocity of 1.0~m/s. Furthermore, we conducted a case study that validates a camera failure mitigation method that is difficult to realize with heavy-duty trucks due to its high risks. We hope our testbed can benefit many researchers by releasing all the details of our work on Github.

In the future, we plan to extend our work by adding more complex safety scenarios such as network failures and computer failures.








\section*{Acknowledgment}
\label{sec:acknowledgement}

This work was supported by Institute of Information and communications Technology Planning and Evaluation (IITP) grant funded by the Korea government (MSIT) (No. 2014-3-00065, Resilient Cyber-Physical Systems Research). J.-C. Kim is the corresponding author of this paper.


\balance
\bibliographystyle{IEEEtran}
\bibliography{platoon}

\end{document}